%% file: paper204.tex
\documentclass[runningheads]{llncs}
\usepackage{graphicx}
\usepackage{amsmath}
\usepackage[labelfont=bf]{caption}
\usepackage{enumitem}  
\usepackage[misc]{ifsym}
\usepackage{graphicx}
\usepackage{threeparttable}
\usepackage{ntheorem}
\usepackage{color}
\usepackage{colortbl}
\usepackage{multirow}
\usepackage{amsmath,amssymb}
\usepackage{comment}
\usepackage{subcaption}
\usepackage{bm}                                            %
\usepackage{etoolbox}                                      
\usepackage{url}
\usepackage{nth}
\usepackage{cite}
\usepackage{balance}
\usepackage[bookmarks=false]{hyperref}                     
\usepackage{courier}                                       
\usepackage{listings}
\usepackage[table,xcdraw]{xcolor}
\usepackage[]{algpseudocode}     
\hypersetup{
    colorlinks = true,
    citecolor  = blue,
    linkcolor  = blue,
    urlcolor   = blue,
}
\definecolor{color1}{HTML}{377eb8}  
\definecolor{co}{HTML}{4daf4a}  
\definecolor{ourscolor}{HTML}{ff7f00}  
\definecolor{gray095}{gray}{0.95}
\definecolor{gray09}{gray}{0.9}
\definecolor{gray085}{gray}{0.85}
\definecolor{gray08}{gray}{0.8}
\definecolor{gray075}{gray}{0.75}
\definecolor{gray07}{gray}{0.7}

\algrenewcommand\textproc{\texttt}
\makeatletter\let\c@float@type\relax\makeatother
\makeatletter\let\float@addtolists\relax\makeatother
\usepackage{algorithm}

\makeatletter
\newcommand{\thickhline}{%
	\noalign {\ifnum 0=`}\fi \hrule height 1pt
	\futurelet \reserved@a \@xhline
}
\DeclareMathOperator*{\argmin}{arg\,min}
\graphicspath{{./figs/}}

\begin{document}

\title{Unsupervised Deformable Image Registration with Structural Nonparametric Smoothing}

\author{%
Hang Zhang\inst{1} \and
Renjiu Hu\inst{1}  \and
Xiang Chen\inst{2} \and
Min Liu\inst{2} \and
Yaonan Wang\inst{2} \and
Rongguang Wang\inst{3} \and
Jinwei Zhang\inst{4} \and
Gaolei Li\inst{5} \and
Xinxing Cheng\inst{6} \and
Jinming Duan\inst{6,7} \textsuperscript{(\Letter)}
}

\authorrunning{H. Zhang et al.}

\institute{%
Cornell University, Ithaca, NY, USA \and 
Hunan University, Changsha, China \and
University of Pennsylvania, Philadelphia, PA, USA \and
Johns Hopkins University, Baltimore, MD, USA \and
Shanghai Jiao Tong University, Shanghai, China \and
University of Birmingham, Birmingham, UK\\ \and
University of Manchester, Manchester, UK \\
\email{Jinming.Duan@manchester.ac.uk}
}

\maketitle 

\input{docs/0_abstract}    
\input{docs/1_intro}

\input{docs/2_related}

\input{docs/3_method}

\input{docs/4_results}

\input{docs/5_conclusion}

\bibliographystyle{splncs04}
\bibliography{paper204}

\end{document}

%% file: docs/0_abstract.tex
\begin{abstract}

Learning-based deformable image registration (DIR) accelerates alignment by amortizing traditional optimization via neural networks.
Label supervision further enhances accuracy, enabling efficient and precise nonlinear alignment of unseen scans.
However, images with sparse features amid large smooth regions, such as retinal vessels, introduce aperture and large-displacement challenges that unsupervised DIR methods struggle to address.
This limitation occurs because neural networks predict deformation fields in a single forward pass, leaving fields unconstrained post-training and shifting the regularization burden entirely to network weights.
To address these issues, we introduce \textbf{SmoothProper}, a plug-and-play neural module enforcing smoothness and promoting message passing within the network's forward pass.
By integrating a duality-based optimization layer with tailored interaction terms, SmoothProper efficiently propagates flow signals across spatial locations, enforces smoothness, and preserves structural consistency.
It is model-agnostic, seamlessly integrates into existing registration frameworks with minimal parameter overhead, and eliminates regularizer hyperparameter tuning.
Preliminary results on a retinal vessel dataset exhibiting aperture and large-displacement challenges demonstrate our method reduces registration error to 1.88 pixels on \(2912 \times 2912\) images, marking the first unsupervised DIR approach to effectively address both challenges.
The source code will be available at \url{https://github.com/tinymilky/SmoothProper}.  

\keywords{Nonparametric smoothing  \and Deformable image registration \and Neural networks \and Plug-and-play \and Aperture problem \and Large displacement}

\end{abstract}

%% file: docs/1_intro.tex
\section{Introduction}
\label{sec:intro}

In recent years, learning-based deformable image registration (DIR) methods, fueled by progress in neural networks \cite{he2016deep,vaswani2017attention,dosovitskiyimage}, have become mainstream for various registration tasks. 
Compared to classical iterative optimization methods \cite{ashburner2007fast,avants2011reproducible,marstal2016simpleelastix,vercauteren2009diffeomorphic,heinrich2013mrf}, learning-based approaches \cite{balakrishnan2019voxelmorph,mok2020large,chen2022transmorph,meng2024correlation,chen2024textscf} achieve faster performance through amortized optimization and offer potentially higher accuracy with the integration of label supervision (e.g., segmentation, keypoints).

\begin{figure}[!t]
    \centering
    \includegraphics[width=0.85\columnwidth]{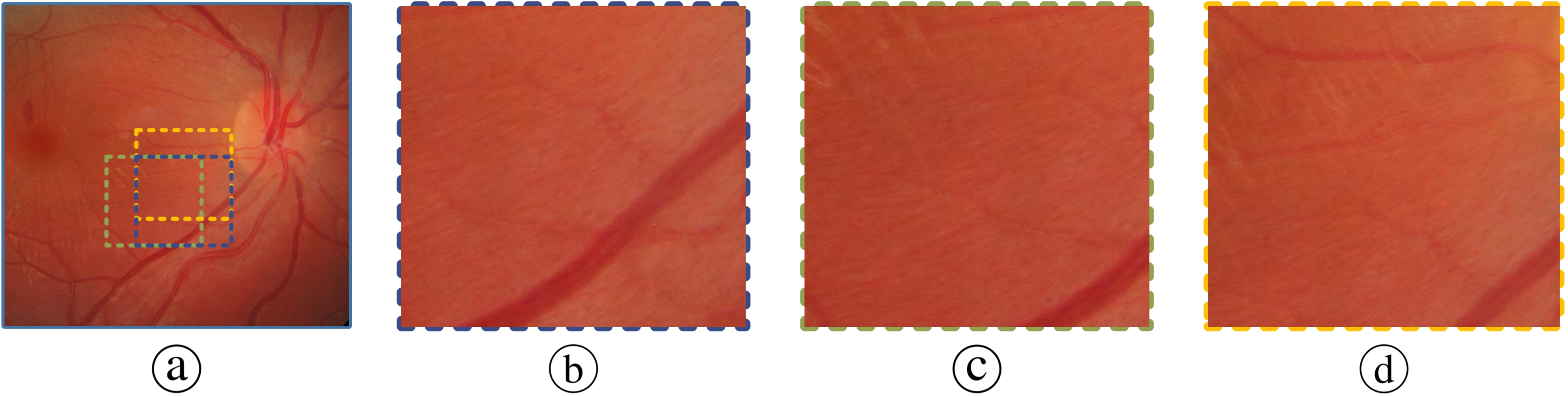}
    \caption{
    Visual illustration of aperture and large displacement challenges. 
    (a) shows a retinal vessel patch from the FIRE dataset, with blue, green, and yellow dashed boxes highlighting local vessel structures at different positions. 
    The green and yellow boxes are horizontal and vertical translations of the blue box, respectively. 
    (b), (c), and (d) are zoomed-in views of the blue, green, and yellow boxes, demonstrating that local information alone is insufficient to reconstruct their displacements. 
    Furthermore, the displacements in (b)-(c) and (b)-(d) exceed the vessel diameter.
    }
    \label{fig:aperture_large_disp}
\end{figure}

Despite promising performance, unsupervised DIR methods struggle with tasks involving both the \textit{aperture problem} and the \textit{large displacement problem}.
The aperture problem arises in homogeneous or textureless regions, where limited local evidence within the network’s \textit{effective receptive field (ERF)} \cite{luo2016understanding} hinders accurate displacement estimation.
The large displacement problem occurs when structural displacement between image pairs exceeds the structure's size.
While many registration tasks involve either issue separately, retinal vessel datasets often present both simultaneously (see Fig.~\ref{fig:aperture_large_disp}), significantly complicating unsupervised DIR.
Current methods primarily rely on descriptor matching approaches \cite{liu2022semi,liu2023geometrized}, and to our knowledge, no existing unsupervised DIR methods have successfully handle both challenges in retinal vessel registration.


\begin{figure}[!t]
    \centering
    \includegraphics[width=0.95\columnwidth]{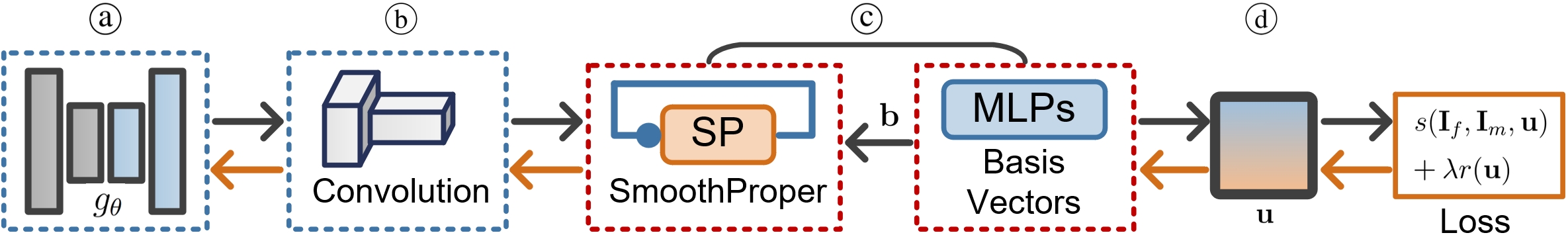}
    \caption{
    Illustration of the registration framework and integration of SmoothProper into learning-based frameworks.
    Black arrows represent the forward pass, while yellow arrows indicate the backward pass.
    Dashed components are optional: removing (a), (b), and (c) reverts to a classic iterative approach where loss gradients directly update the flow field \(\mathbf{u}\), similar to ConvexAdam \cite{siebert2024convexadam}, which follows a modern learning framework but aligns with this strategy.
    Removing (c) leads to a conventional learning-based approach, where the \(r(\mathbf{u})\) affects the flow field only through network weights via amortized optimization. 
    Most learning-based frameworks, including pyramid or multi-scale models (except unrolling methods \cite{heinrich2019closing,qiu2022embedding,jia2021learning}), use a backbone network followed by convolution (b) to generate the flow field. 
    Removing (b) allows SmoothProper with a basis vector generator to replace this convolution layer, adding minimal parameters.
    }
    \label{fig:reg_framework}
\end{figure}

We hypothesize that the impact of the pairwise regularization term in the loss function has been underestimated. 
As demonstrated by the Horn-Schunck optical flow equation \cite{horn1981determining}, this regularizer not only smooths the flow field but also facilitates message passing \cite{heinrich2019closing,hansen2021revisiting,hansen2021deep,zhang2024slicer}, propagating flow information from regions with strong signals (e.g., boundaries, texture-rich areas) to those with ambiguous or missing signals (e.g., large homogeneous areas).
This mechanism is essential for addressing the aperture and large displacement problems. 
However, most unsupervised DIR methods incorporate this regularizer only as a loss term, shifting the burden for smoothness and signal propagation entirely to the network weights. 
See Fig. \ref{fig:reg_framework} for a visual illustration.

To address these challenges, we introduce \textbf{SmoothProper}, a plug-and-play unrolled neural network layer seamlessly integrating into existing frameworks with minimal parameter overhead (see Fig.~\ref{fig:reg_framework}, removing (b) or (c) for illustration).
Our approach extends the Horn-Schunck variational formulation \cite{horn1981determining}, employing a duality-based optimization framework \cite{rudin1992nonlinear,chambolle2004algorithm,zach2007duality} with quadratic relaxation \cite{steinbrucker2009large} to decouple data fidelity from smoothness regularization.
Unlike prior unrolling methods \cite{qiu2022embedding,jia2021learning,blendowski2021weakly,hu2024a} emphasizing raw data consistency or neural network-based denoising, SmoothProper introduces adaptive, conditional basis vectors and tailored interaction terms in each unrolled iteration.
This structure enables message passing, preserves local structural consistency, and eliminates the need for tuning regularization hyperparameters.
SmoothProper resembles MoDL's unrolled architecture \cite{aggarwal2018modl} and incorporates mechanisms analogous to loopy belief propagation \cite{felzenszwalb2006efficient} and mean-field message passing \cite{krahenbuhl2011efficient}.
The main contributions of this work are as follows:
\begin{itemize}
    \item We introduce SmoothProper, an unrolled neural network layer that enforces smoothness and message passing with local structural consistency during the network forward pass, addressing the aperture and large-displacement challenges in unsupervised DIR.
    \item SmoothProper is lightweight, model-agnostic, integrates seamlessly into unsupervised DIR frameworks with minimal parameter overhead, and removes the need for regularizer hyperparameter tuning.  
    \item We have achieved the first unsupervised learning-based retinal vessel registration, reducing the average error to 1.88 pixels on \(2912 \times 2912\) images.  
\end{itemize}

%% file: docs/2_related.tex
\section{Related Work}

\textbf{Preliminaries on Learning-based DIR Methods:}
Classical iterative methods calculate a flow field by solving a \textit{pairwise} variational optimization problem \cite{horn1981determining}. 
This allows direct updates to the flow field \(\mathbf{u}\) using loss gradients (see Fig. \ref{fig:reg_framework} for a visual example). 
In contrast, learning-based methods employ \textit{amortized optimization} over network parameters \(\theta\) across a cohort of image pairs $D$:
\begin{equation}
\min_{\theta} ~ \mathbb{E}_{(\mathbf{I}_f,\mathbf{I}_m) \sim D} [s(\mathbf{I}_f, \mathbf{I}_m, \mathbf{u}) + \lambda r(\mathbf{u})], ~~ \text{s.t.} ~~ \mathbf{u} = g_{\theta}(\mathbf{I}_f,\mathbf{I}_m).
\label{eq:horn_schunck}
\end{equation}
Here, \(\mathbf{I}_f\) and \(\mathbf{I}_m\) denote the fixed and moving images, respectively; \(\mathbf{u}\) is the displacement field output by a neural network \(g\) parameterized by \(\theta\); \(\lambda\) is a hyperparameter; \(s\) is the dissimilarity function; and \(r\) is typically a pairwise diffusive regularizer.
Training can be accomplished by accumulating gradient backflows with respect to \(\theta\):
\begin{equation}
    \frac{\partial \mathcal{L}}{\partial \theta} = \sum_{f,m} \frac{\partial s}{\partial g_{\theta}} \frac{\partial g_{\theta}}{\partial \theta} + \lambda \sum_{f,m} \frac{\partial r}{\partial g_{\theta}} \frac{\partial g_{\theta}}{\partial \theta}
    \label{eq:theta_gradient}
\end{equation}
Here, \(\mathcal{L}\) denotes the total loss, with the Jacobians mapping flow gradients to network weights. 
As noted by Hansen and Heinrich \cite{hansen2021revisiting}, and shown in recent work \cite{jena2024deep}, unsupervised DIR methods relying on amortized optimization yield limited gains in label matching (e.g., Dice score), since the first Jacobian in Eq.~\eqref{eq:theta_gradient} contains no more information than raw intensity-label mutual information. 
We hypothesize that while the first Jacobian fails to improve label matching, the second Jacobian, beyond enforcing smoothness, can function as an effective message passing mechanism \cite{felzenszwalb2006efficient,hansen2021revisiting} if properly utilized. 
This potential has been largely overlooked, and our SmoothProper addresses it via a cost-efficient, easily integrated architecture grounded in modern neural network design.

\textbf{Learning-based and Deep Unrolling DIR Methods:}
To address aperture and large-displacement challenges, increasing the network ERF is common, involving methods like transformers \cite{vaswani2017attention,chen2022transmorph,shi2022xmorpher}, large-kernel convolutions \cite{ding2021repvgg,jia2022u}, Laplacian pyramids \cite{mok2021large,wang2024recursive,memwarp2024}, and coarse-to-fine designs \cite{meng2024correlation,ma2024iirp}.  
However, relying solely on the loss term and network weights for message passing may risk the network acting as a low-pass filter \cite{rahaman2019spectral,wang2022anti}, leading to oversmoothing that obscures structural details.  
To overcome these limitations, deep unrolling \cite{hershey2014deep} integrates model-based methods \cite{aggarwal2018modl,zhang2020fidelity,zhang2023deda,zhang2023laro} and iterative algorithms \cite{sun2016deep,chen2018theoretical,gregor2010learning} directly into neural network architectures.  
Recent studies \cite{heinrich2019closing,jia2021learning,blendowski2021weakly,qiu2022embedding,hu2024a} have applied deep unrolling to image registration, enforcing both dissimilarity and regularity functions during the network forward pass.  
Among these works, most focus on either raw data consistency or neural network denoisers as regularizers, while PDD-Net \cite{heinrich2019closing} remains the only one that explicitly enforces smoothness, though it relies on a costly 6D dissimilarity map and underutilizes neural network capacity for message passing.
GraDIRN \cite{qiu2022embedding} uses unrolled gradient descent for data consistency, PRIVATE \cite{hu2024a} employs a plug-and-play denoiser with instance-wise optimization, and SUITS \cite{blendowski2021weakly} and VR-Net \cite{jia2021learning} rely on separate feature extraction or alternating data consistency and denoising steps.
In contrast, SmoothProper directly integrates smoothness regularization into the network forward pass, enabling efficient message passing without label supervision or heavy computational overhead.


%% file: docs/3_method.tex
\section{Method}


This section introduces the bi-level optimization framework, defines SmoothProper and its optimization, presents a conditional automatic adjustment for \(\lambda\), and outlines the overall framework integrating SmoothProper.  


\subsection{The SmoothProper}

\textbf{Bi-level Optimization framework:}
The proposed SmoothProper adds a sub-optimization problem as a constraint to Eq.~\eqref{eq:horn_schunck}:
\begin{subequations} \label{eq:bi_level}
\begin{align}
    \min_{\theta} ~ & \mathbb{E}_{(\mathbf{I}_f,\mathbf{I}_m) \sim D} [\mathcal{L}(\mathbf{I}_f, \mathbf{I}_m, \mathbf{u}^{*}(\theta))], \label{eq:bi_level_a} \\
    \text{s.t.} ~ & \mathbf{u}^{*}(\theta) = \argmin_{\mathbf{u}} \mathcal{S}(\mathbf{I}_f,\mathbf{I}_m,\mathbf{u},\theta). \label{eq:bi_level_b}
\end{align}
\end{subequations}
Eq.~\eqref{eq:bi_level_b} constrains the expected loss in Eq.~\eqref{eq:bi_level_a}, optimizing the displacement field by minimizing the SmoothProper function \(\mathcal{S}\) based on input images and network parameters. 
The optimization \(\mathcal{S}\) is embedded in the registration network by unrolling its iterative steps into the forward pass, with trainable parameters \(\theta\). 
Given the solution \(\mathbf{u}^*\) for each image pair in the training set \(D\), Eq.~\eqref{eq:bi_level_a} minimizes the expected loss \(\mathcal{L}(\mathbf{I}_f, \mathbf{I}_m, \mathbf{u}^*(\theta))\) to optimize \(\theta\). 
Unlike GraDIRN \cite{qiu2022embedding}, which unrolls its energy function to enforce data consistency, our approach unrolls \(\mathcal{S}\) to shift the regularization burden away from network weights.

\textbf{Quadratic Relaxation:}
Following prior works \cite{chambolle2004algorithm,zach2007duality,vercauteren2009diffeomorphic,steinbrucker2009large}, we adopt a quadratic relaxation of Eq.~\eqref{eq:horn_schunck} to decouple the data consistency term and the regularizer:
\begin{equation}
    \min_{\mathbf{u}} \int_{\Omega} \left( \rho(\mathbf{v}) + \frac{1}{2\alpha} \|\mathbf{v}-\mathbf{u}\|^2 + \beta r(\mathbf{u}) \right) \, \mathrm{d}\Omega.
    \label{eq:quad_relax}
\end{equation}
Here, \(\rho(\mathbf{v})\) measures the cost of applying the flow field \(\mathbf{v}\) to the input image pair, while \(\beta r(\mathbf{u})\) represents the smoothness regularizer with strength \(\beta\).
It has been shown \cite{chambolle2004algorithm} that as \(\alpha \rightarrow 0\), minimizing Eq.~\eqref{eq:quad_relax} with the same data and regularization terms is equivalent to minimizing it without the interaction term.
Although non-quadratic relaxation terms exist \cite{black1993framework}, we empirically found the suboptimization problem easier to solve, whereas using an L1 term requires proximal gradients, adding complexity.

Many gradient-descent methods rely on first-order Taylor expansions to linearize \(\rho(\mathbf{v})\), limiting them to small steps and requiring numerous iterations, which hinders handling large deformations. Deep unrolled DIR methods \cite{qiu2022embedding,blendowski2021weakly,jia2021learning}, built on gradient-based optimization, inherit these limitations. 
In contrast, discrete optimization approaches \cite{heinrich2013mrf,heinrich2014non,siebert2024convexadam} rely on predefined cost volumes, allowing each subproblem for \(\mathbf{v}\) and \(\mathbf{u}\) to be convex and optimally solvable, thus better accommodating large deformations. 
However, their computational cost rises sharply when expanding the discrete space to cover larger displacements.

\textbf{Objective Function:}
SmoothProper leverages neural networks to adaptively linearize local intensities using a basis-constrained formulation.  
We discretize Eq.~\eqref{eq:quad_relax} as a sum of voxel- and pair-wise energies, with the basis matrix \(\mathbf{b} \in \mathbb{R}^{m \times d}\), where \(\mathbf{b} = [\mathbf{b}_1, \mathbf{b}_2, \dots, \mathbf{b}_m]^{\top}\) and each \(\mathbf{b}_i \in \mathbb{R}^{d}\) is a basis vector (\(d = 2\) for 2D images).  
Given the neural network output \(\mathbf{p} = g_{\theta}(\mathbf{I}_f, \mathbf{I}_m)\), where \(\mathbf{p}(x) \in \mathbb{R}^{m}\) is a non-negative coefficient vector (from ReLU), representing the flow signal strength at \(x \in \Omega\), the basis vectors are linearly combined to produce the displacement fields.  
This leads to the following energy optimization:
\begin{subequations} \label{eq:smooth_proper}
\begin{align}
   \min_{\mathbf{q},\mathbf{v}} &\sum_{x \in \Omega}\|\mathbf{p}(x)-\mathbf{q}(x)\|^2 + \sum_{x \in \Omega} \sum_{i=1}^{m}\frac{1}{2\alpha}\mathbf{q}_i(x) \|\mathbf{v}(x)-\mathbf{b}_i\|^2
   \nonumber\\
    +&\sum_{x \in \Omega} \frac{1}{2\alpha} \|\mathbf{q}(x)\mathbf{b}-\mathbf{v}(x)\|^2 +\beta\mathbf{r}(\mathbf{v}). \tag{5}
\end{align}
\end{subequations}
Here, \(\mathbf{q}(x)\mathbf{b}\) computes the displacement at location \(x\), while \(\mathbf{v}(x)\) represents the auxiliary displacement vector at \(x\). 
The first term aligns \(\mathbf{q}(x)\) with the network’s initial prediction \(\mathbf{p}(x)\), while the fourth term \(\mathbf{r}(\mathbf{v}) = \|\nabla \mathbf{v}\|^2\) is a diffusive regularizer. 
The second term, unique to this work, introduces a directional bias between \(\mathbf{q}(x)\) and \(\mathbf{b}_i\), ensuring the displacement can be closely represented by a linear combination of the basis vectors \(\mathbf{b}_i\), thereby guiding the reconstruction of displacement field from $\mathbf{q}$. 
The third coupling term pulls \(\mathbf{v}(x)\) toward the linear combination \(\mathbf{q}(x)\mathbf{b}\).
This creates a feedback loop that enables two key properties: 
\begin{itemize}
    \item \textbf{Message Passing}: The term \(\|\nabla \mathbf{v}\|^2\) encourages gradual changes in \(\mathbf{v}\), smoothing \(\mathbf{q}\) as \(\mathbf{q}\mathbf{b}\) aligns with \(\mathbf{v}\). This results in \(\mathbf{q}\) serving as a smooth approximation of \(\mathbf{p}\), effectively passing flow signals across regions.
    \item \textbf{Structural Consistency}: As \(\mathbf{q}(x)\) and \(\mathbf{v}(x)\) iteratively align with \(\mathbf{p}(x)\) and the basis \(\mathbf{b}_i\), the equation ensures that regions with strong flow signals (large \(\mathbf{q}(x)\)) remain anchored to their representative basis patterns $\mathbf{b}_i$. 
\end{itemize}
This basis-constrained formulation employs a “\textbf{smooth-reinforce}” interplay:  
The smoothing step diffuses signals, slightly diminishing strong ones and enhancing weaker ones.  
The reinforcement step then “\textbf{locks in}” strong signals: when \(\mathbf{p}(x)\) is large, the second term keeps \(\mathbf{v}(x)\) close to the corresponding basis pattern, preventing dilution.  
Meanwhile, weak signals, associated with smaller \(\mathbf{q}(x)\),can be influenced by stronger signals spreading through message passing of \(\mathbf{p}(x)\). 
This iterative smooth-reinforce cycle promotes signal propagation across regions, preserves strong signals, and effectively extends them into weaker areas, addressing the aperture problem and mitigating oversmoothing.


Unlike previous methods that rely on manual linearization \cite{jia2021learning,blendowski2021weakly} or omit it entirely \cite{hu2024a,qiu2022embedding}, our proposed Eq.~\eqref{eq:smooth_proper} adaptively linearizes image intensities within the network’s effective receptive field (ERF) into learned coefficients.  
In contrast to PDD-Net \cite{heinrich2019closing}, which uses fixed displacement vectors and pre-computed cost volumes, SmoothProper jointly learns both basis vectors and coefficients during training.  
These learned basis vectors serve as “dictionary atoms” \cite{sukhbaatar2015end,memwarp2024} or clustering centers \cite{van2017neural,esser2021taming}, providing an adaptive representation that directly influences field smoothness.  
This approach achieves higher accuracy without computing a full cost volume, instead encoding disparity context directly within the ERF.  
Figure~\ref{fig:sp_recursive} illustrates the recursive SmoothProper with its feedback loop.

\subsubsection{Unrolling the Recursive Networks.}
\label{sec:unroll_rnn}
\begin{figure}[!t]
\centering
\captionsetup[sub]{font=scriptsize}
    \subcaptionbox{
    The SmoothProper framework.
    \label{fig:smooth_proper_framework}
    }{
        \includegraphics[width=0.47\columnwidth]{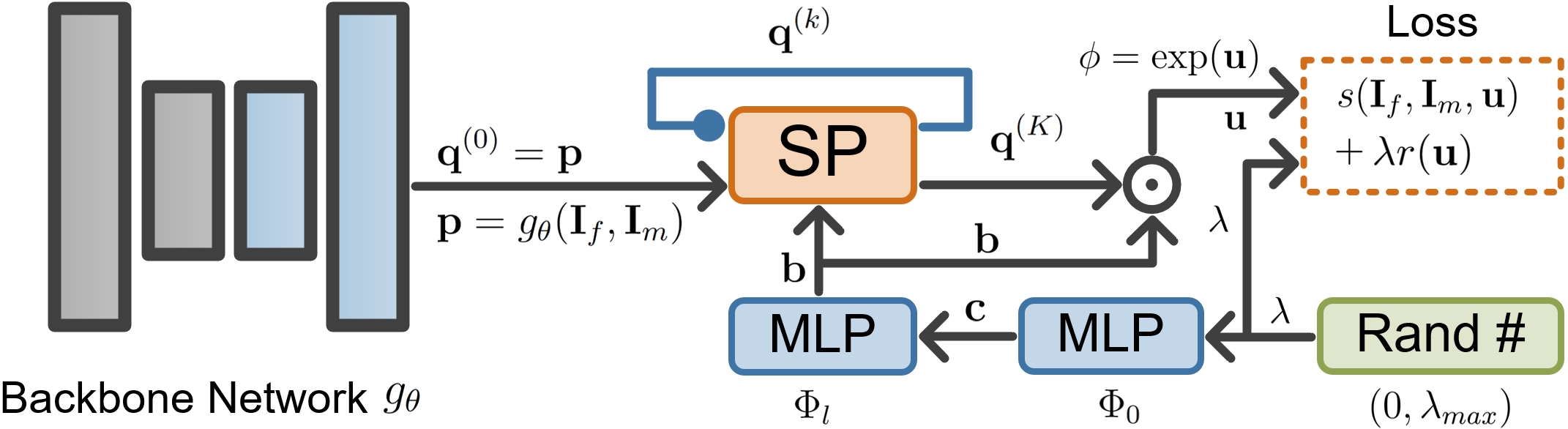}
    }
    \subcaptionbox{
    The Recursive SmoothProper.
    \label{fig:sp_recursive}
    }{
        \includegraphics[width=0.47\columnwidth]{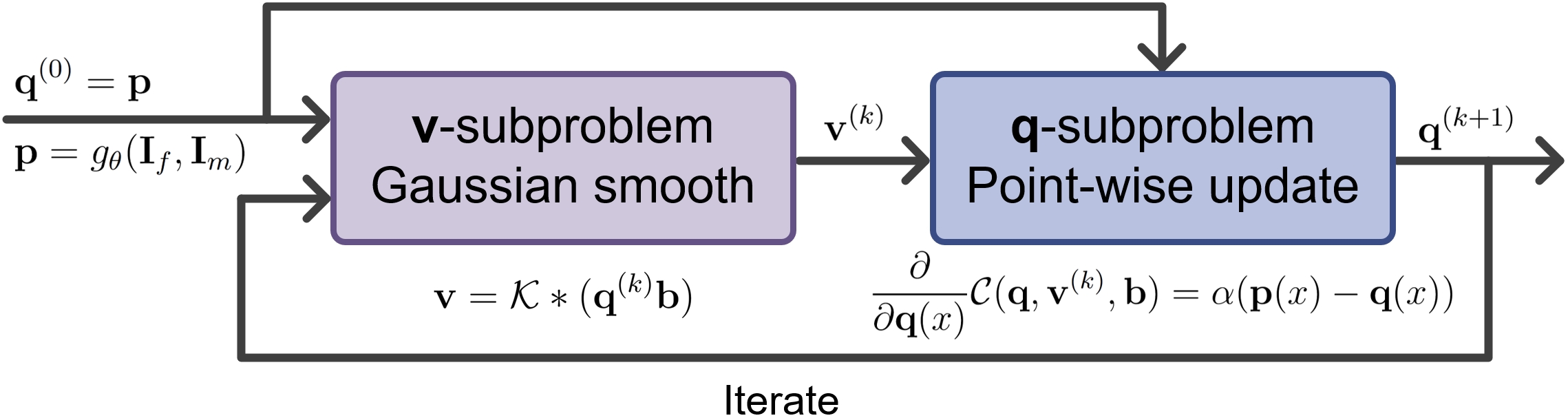}
    } \\
    \vspace{2.5ex}
    \subcaptionbox{
    The unrolled SmoothProper as described in Eq.~\eqref{eq:unrolled_sp} 
    \label{fig:sp_unroll}
    }{
        \includegraphics[width=1.\columnwidth]{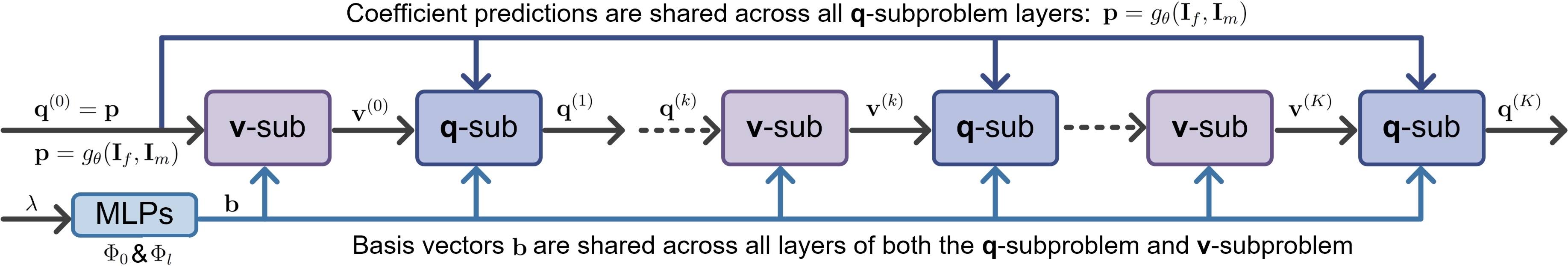}
    }
    \caption{
    (a) is the schematic of the SmoothProper (SP) framework. 
    The yellow box represents SP optimization, where \(\mathbf{p} = g_{\theta}(\mathbf{I}_f, \mathbf{I}_m)\) and \(\mathbf{b} = \Phi_l(\Phi_0(\lambda))\), with \(\lambda\) randomly sampled from \((0, \lambda_{max})\). 
    SP iteratively updates \(\mathbf{q}^k\) to \(\mathbf{q}^{k+1}\) until the maximum iteration count \(K\) is reached. 
    (b) and (c) are visualizations of the SmoothProper module architecture in recursive and unrolled forms. 
    Basis vectors \(\mathbf{b}\) and \(\alpha\) are omitted for brevity (details in \S\ref{sec:unroll_rnn}).
    }
    \label{fig:sp_rnn_unroll}
\end{figure}

To solve Eq.~\eqref{eq:smooth_proper}, we employ coordinate descent, alternating between \(\mathbf{q}^{(k)}\) and \(\mathbf{v}^{(k)}\) while progressively reducing \(\alpha\) to achieve \(\mathbf{q}^{(K)}\mathbf{b} \approx \mathbf{v}^{(K)}\), thus approximating the solution of Eq.~\eqref{eq:smooth_proper}. Here, \(K\) is the total number of iterations, and \(k\) denotes the \(k_{th}\) iteration. 
Following prior work \cite{heinrich2014non}, we set \(\alpha = \{150, 50, 15, 5, 1.5, 0.5\}\) for \(K = 6\) alternating iterations.
Let the directional bias and coupling terms be represented by \(\mathcal{C}(\mathbf{q}, \mathbf{v}, \mathbf{b})\), we decompose the optimization of Eq.~\eqref{eq:smooth_proper} into two subproblems as follows:
\begin{subequations} \label{eq:unrolled_sp} 
\begin{align} \mathbf{q}^{(k)} = \argmin_{\mathbf{q}} &~ \frac{1}{2\alpha}\mathcal{C}(\mathbf{q}, \mathbf{v}^{(k-1)}, \mathbf{b}) + \sum_{x \in \Omega} \|\mathbf{p}(x) - \mathbf{q}(x)\|^2; \nonumber \\ \mathbf{v}^{(k)} = \argmin_{\mathbf{v}} &~ \frac{1}{2\alpha}\mathcal{C}(\mathbf{q}^{(k)}, \mathbf{v}, \mathbf{b}) + \beta \|\nabla \mathbf{v}\|^2. \tag{6}
\end{align} 
\end{subequations}
The initial \(\mathbf{q}^{(0)}\) is set to \(\mathbf{p} = g_{\theta}(\mathbf{I}_f, \mathbf{I}_m)\), the output from the backbone network \(g_{\theta}\). 
The \(\mathbf{q}\) subproblem can be solved optimally and pointwise across \(x \in \Omega\) in closed form by taking the derivative of \(\mathcal{C}(\mathbf{q}, \mathbf{v}^{(k-1)}, \mathbf{b})\) with respect to \(\mathbf{q}(x)\) and setting it to zero:
\begin{equation}
\frac{\partial}{\partial \mathbf{q}(x)} \mathcal{C}(\mathbf{q}, \mathbf{v}^{(k-1)}, \mathbf{b}) = \alpha \big(\mathbf{p}(x) - \mathbf{q}(x)\big).
\label{eq:q_sub}
\end{equation}
The \(\mathbf{v}\) subproblem is solved through fixed-point iteration \cite{chambolle2004algorithm}, involving neighboring values of \(\mathbf{v}(x)\) and \(\mathbf{q}(x)\mathbf{b}\), yielding \(\mathbf{v}\) as a smoothed version of \(\mathbf{q}\mathbf{b}\). 
This can be approximated by applying Gaussian blurring to \(\mathbf{q}\mathbf{b}\) as \(\mathbf{v}^{(k)}=\mathcal{K} * \mathbf{q}^{(k)}\mathbf{b}\) across spatial locations, with the blurring strength determined by \(\beta\). 
A similar approach for optimizing \(\mathbf{v}\) is discussed in Diffeomorphic Demons \cite{vercauteren2009diffeomorphic}. 
The unrolled SmoothProper architecture is illustrated in Fig.~\ref{fig:sp_unroll}.

\subsection{Conditional Learning of Basis Vectors}
SmoothProper serves as the constraint in Eq.~\eqref{eq:bi_level}, providing a smoothed version of \(\mathbf{p} = g_{\theta}(\mathbf{I}_f, \mathbf{I}_m)\) with \(\mathbf{q}\). 
The learnable basis vectors \(\mathbf{b}\) link the smoothness of \(\mathbf{v}\) to \(\mathbf{q}\), adjusting the weight of \(\mathbf{q}\) based on its proximity to \(\mathbf{v}\). 
This enables adaptive regularization strength in the outer loop of Eq.~\eqref{eq:bi_level_a} through \(\mathbf{b}\). 
We use two MLPs, \(\Phi_0\) and \(\Phi_l\): \( \mathbf{c}\leftarrow\Phi_0(\lambda)\) generates a latent code \(\mathbf{c}\), and \(\mathbf{b}\leftarrow\Phi_l(\mathbf{c})\) produces conditional basis vectors. 
Using two MLPs instead of one provides model-agnostic flexibility, allowing \(\mathbf{c}\) to be shared across all pyramid or scale levels if the model is multi-scale-based, while each level has its own \(\Phi_l\).

During training, we randomly sample \(\lambda \in (0, \lambda_{max})\) at each forward pass, enabling \(\mathbf{b}\) and the from-matching strength \(\lambda\) to adapt via \(\Phi_l(\Phi_0(\lambda))\).  
This eliminates the need for manual \(\lambda\)-tuning and allows smoothness strength adjustments after amortized optimization.
Prior works (e.g., HyperMorph \cite{hoopes2021hypermorph} and cLapIRN \cite{mok2021conditional}) also use hypernetworks \cite{ha2017hypernetworks}, but they rely on inefficient, non-scalable designs: HyperMorph conditions the entire network, and cLapIRN requires extensive architectural changes.
In contrast, our method is lightweight and efficient, requiring only a single convolution layer replacement with a SmoothProper layer at the backbone’s end, leaving the backbone unchanged (Fig.~\ref{fig:reg_framework}).

\subsection{Overall Framework}
Fig.~\ref{fig:smooth_proper_framework} shows the SmoothProper framework, which consists of three components: a backbone network \(g_{\theta}\) for initial coefficient estimation, MLPs \(\Phi_0\) and \(\Phi_l\) for latent code and conditional basis vector generation, and unrolled SmoothProper layer.
The SmoothProper module processes the network-predicted coefficients \(\mathbf{p}\) and the MLP-predicted basis vectors \(\mathbf{b}\) through \(K\) iterations to produce \(\mathbf{q}^{(K)}\), a smooth and structurally consistent version of \(\mathbf{p}\). 
The flow field \(\mathbf{u}\) is obtained by linearly combining \(\mathbf{q}^{(K)}\) with \(\mathbf{b}\) and is integrated using scaling and squaring \cite{arsigny2006log} to generate a diffeomorphic deformation field \(\phi\). 
As in prior work \cite{dalca2019unsupervised}, the number of integration steps is fixed to 7. 
The deformation field \(\phi\) is used for the dissimilarity term, while \(\mathbf{u}\) is used for the regularization term in the loss function. 
In summary, SmoothProper not only smooths the displacement field \(\mathbf{u}\) through the diffusive regularizer but also enables message passing, where flow signals (represented by coefficient vectors \(\mathbf{p}\)) propagate with local structural consistency preserved via the feedback loop formed by the interaction term \(\mathcal{C}(\mathbf{q}, \mathbf{v}, \mathbf{b})\).

%% file: docs/4_results.tex
\section{Experiments \& Results}

\subsection{Datasets \& Evaluation Metrics}

We evaluated our method on the public Fundus Image Registration (FIRE) dataset \cite{hernandez2017fire}, which contains 134 image pairs from 39 subjects. 
Captured at \(2912 \times 2912\) resolution with a \(45^\circ\) field of view using a Nidek AFC-210 fundus camera, the images were collected at Papageorgiou Hospital, Aristotle University of Thessaloniki. 
Each pair includes 10 annotated landmarks. 
To enhance details, we extract the green channel (richest in structural information due to red-free photography), apply Contrast Limited Adaptive Histogram Equalization (CLAHE) for local contrast enhancement, and use gamma correction (\(\gamma = 1.2\)) for better visibility. 
The dataset is divided into three categories: \(\mathcal{S}\) (71 pairs, $>75\%$ overlap, no anatomical changes), \(\mathcal{P}\) (49 pairs, $<75\%$ overlap, no anatomical changes), and \(\mathcal{A}\) (14 pairs, $>75\%$ overlap, anatomical changes). 
It was stratified by category and split into train, val, and test sets in a 7:1:2 ratio.

We evaluated model performance using Target Registration Error (TRE) and Area Under the Curve (AUC) for TRE thresholds.  
TRE measures the L2 distance between corresponding points in the fixed and warped images, with lower values indicating better alignment.  
AUC (AUC@15, AUC@25, AUC@50) represents the percentage of TRE values below specific thresholds, providing a normalized measure of registration success.  
We also report multiply-add operations (G) and parameter size (MB) to assess model complexity.

\subsection{Baseline Methods \& Implementation Details}
For retinal vessel datasets, most studies focus on descriptor matching methods using homography transformations. 
We benchmarked both detector-based methods, including SuperPoint \cite{detone2018superpoint}, Glampoints \cite{truong2019glampoints}, R2D2 \cite{revaud2019r2d2}, SuperRetina \cite{liu2022semi}, LightGlue \cite{lindenberger2023lightglue} with DISK \cite{tyszkiewicz2020disk}, SiLK \cite{gleize2023silk}, XFeat \cite{potje2024xfeat}, and RetinaIPA \cite{wang2024retinal}, and detector-free methods, such as LoFTR \cite{sun2021loftr}, DKM \cite{edstedt2023dkm}, AspanFormer \cite{chen2022aspanformer}, RoMa \cite{edstedt2023roma}, and GeoFormer \cite{liu2023geometrized}.  
We also evaluated learning-based methods, including KeyMorph \cite{wang2023robust} and C2FViT \cite{mok2022affine}, which claim to address large misalignments and arbitrary rotations (\([-180^\circ, 180^\circ]\)). 
Additionally, we included deep unrolling methods, such as VR-Net \cite{jia2021learning}, GraDIRN \cite{qiu2022embedding}, and PDD-Net \cite{heinrich2019closing}, as well as GAMorph \cite{liu2024progressive}, the only known method co-training descriptor matching with deformable registration, which shows promising results.
\input{docs/quant_fire}

All experiments, including SmoothProper and baseline models, were conducted on a machine with an NVIDIA A100 GPU (80GB), a 16-core CPU, and 32GB RAM, using Python 3.7 and PyTorch 1.9.0.
We used the Adam optimizer \cite{kingma2014adam} with a polynomially decayed learning rate starting at \(4 \times 10^{-4}\), a batch size of 1, 100 epochs, and a regularization strength \(\lambda = 1\) in Eq.~\eqref{eq:horn_schunck} for all learning-based methods across both datasets.
Local normalized cross-correlation served as the dissimilarity metric, with $r(\mathbf{u})=\|\nabla \mathbf{u}\|^2$ as the regularirizer for learning based methods as well as outer loop regularizer for SmoothProper.
For descriptor matching methods, only GeoFormer \cite{liu2023geometrized}, SuperRetina \cite{liu2022semi}, and RetinaIPA \cite{wang2024retinal}, designed specifically for retinal vessel registration, were trained from scratch on the FIRE dataset. 
For other models, fine-tuning on FIRE resulted in poorer performance compared to their pre-trained weights, likely due to the dataset’s small size.
For all learning-based methods, we initialized alignment using the best-performing descriptor matching method. 
This initial alignment is crucial for unsupervised DIR, as extreme conditions (e.g., 180° rotation) may surpass the capability of deformable-based methods.  

While SmoothProper is model-agnostic, we employ a simple three-level Laplacian image pyramid as its backbone network.
It uses three 2D convoloution, instance normalization and ReLU activations blocks (Conv-Block) as encoder to extract moving and fixed feature maps, followed by bilinear downsampling to form a 3-level image pyramid. 
At each pyramid level, the fixed feature map along with the moving feature map wapred by the previous level's deformation field (if any) concatenated through channels, followed by three 2D Conv-Blocks for coefficient vector extraction, and finalized with our SmoothProper module to produce the final field.
The result of SmoothProper in Table~\ref{table:quant_fire} uses the parameters \(m = 4 \times 3^2\), \(K = 6\), and an input image size of \(1024 \times 1024\).  
The effects of these parameters are discussed in \S\ref{sec:results} and shown in Fig.~\ref{fig:ablation_fire}.

\subsection{Results \& Analysis}
\label{sec:results}

\begin{figure}[!t]
    \centering
    \includegraphics[width=0.85\linewidth]{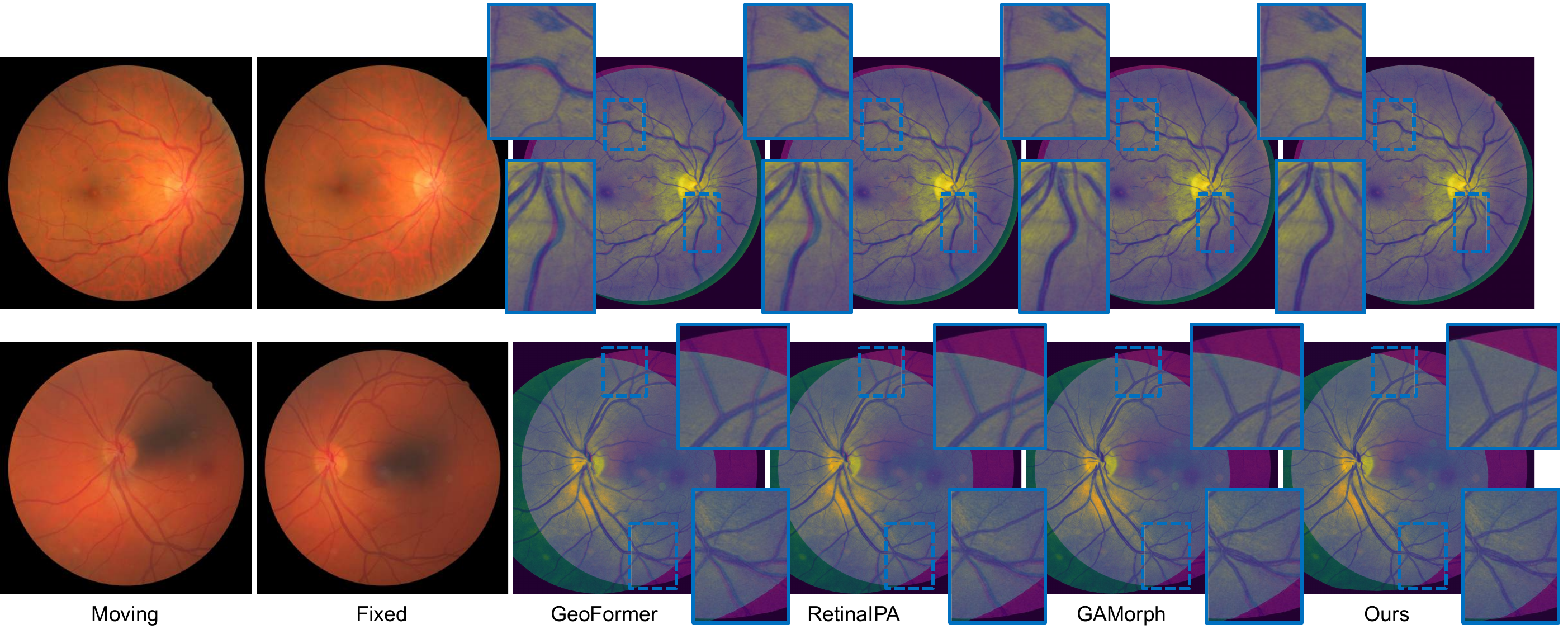}
    \caption{
    Qualitative comparison.
    Fixed images are shown with the 'inferno' colormap, and moving images are overlaid with the 'viridis' colormap for clarity.  
    Highlighted shadow regions show challenging cases with local anatomical changes, geometric shifts, or image shadowing.    
    }
    \label{fig:quant_fire}
\end{figure}

\noindent \textbf{Quantitative Results:}  
Table \ref{table:quant_fire} presents the quantitative results on the FIRE dataset, where the proposed SmoothProper outperforms all other methods across all metrics.  
This marks the first successful application of unsupervised DIR for retinal vessel registration, achieving a TRE of 1.88 pixels on \(2912 \times 2912\) images.  
Specifically, SmoothProper reduces TRE by 67.32\% over RetinaIPA (detector-based), 69.68\% over GeoFormer (detector-free), 39.01\% over GAMorph (learning-based w/o unrolling), and 62.20\% over VR-Net (learning-based w/ unrolling).  
SmoothProper also achieves the highest mAUC@25, the key metric for validating registrations with TRE within 25 pixels.  
Notably, in the breakdown of mAUC@25 into categories \(\mathcal{A}\), \(\mathcal{P}\), and \(\mathcal{S}\), learning-based methods, especially GAMorph and SmoothProper, outperform descriptor-based methods significantly.  
This reflects the limitations of homography-based descriptor matching in longitudinal retinal imaging, where local deformations occur over time.
In addition, most learning-based DIR methods, except GAMorph and VR-Net, perform worse than their initial alignment from descriptor matching method RetinalIPA.  
This likely occurs because the vessel regions occupy a small portion of the image, causing loss-term-based smooth regularization or message passing to dilute vessel displacements within large, smooth, signal-less regions.

\begin{figure}[!t]
    \centering
    \includegraphics[width=0.85\linewidth]{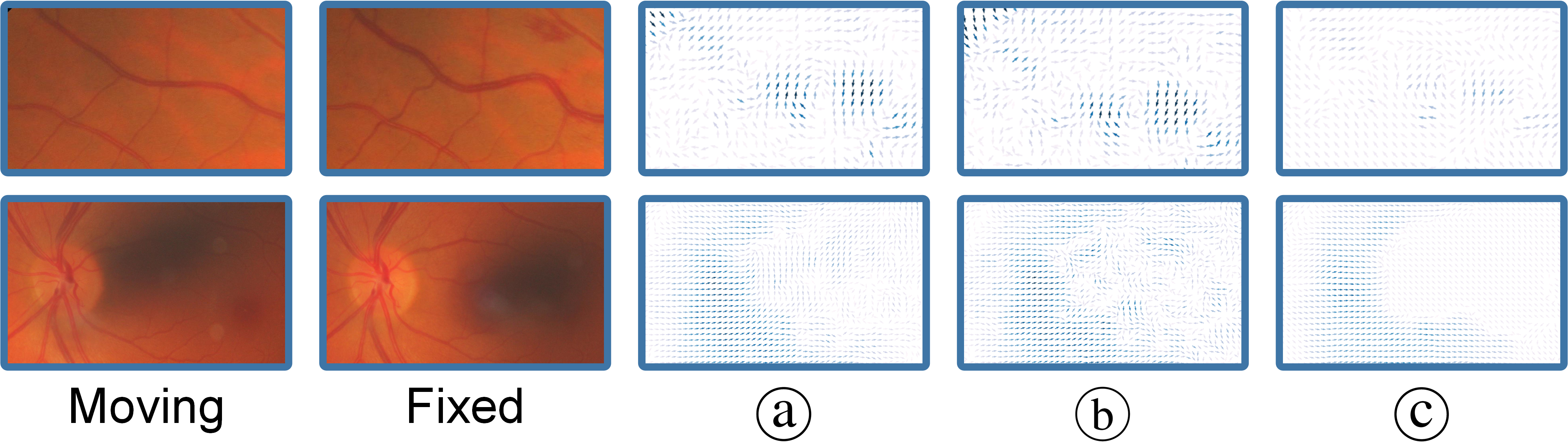}
    \caption{Qualitative comparison of deformation fields: (a) Original SmoothProper, (b) SmoothProper without conditional basis vectors, and (c) GAMorph.
    Darker displacement vectors indicate larger magnitudes.}
    \label{fig:deformation_fields}
\end{figure}

\textbf{Qualitative Results:}  
Fig.~\ref{fig:quant_fire} shows the discrepancy between warped moving images and fixed images for exemplar methods in each category.  
Descriptor matching methods struggle with local deformations, and GAMorph shows limited adaptability.  
In contrast, SmoothProper best adapts to local deformations and is less affected by shadow artifacts. 
Fig.~\ref{fig:deformation_fields} presents local image patches and their displacement fields.  
Comparing (a) and (c), SmoothProper better preserves local vessel structures, avoids shadowing artifacts, and maintains tissue integrity with a regular field that prevents oversmoothing or over-deemphasis.

\textbf{Complexity and Smoothness Analysis:} 
The SmoothProper (SP) counterpart without SP and CBV has 0.32 MB parameters and 31.84 G multi-adds. 
With SP (no CBV), it uses 0.32 MB and 31.95 G multi-adds.  
With SP and CBV, it increases to 41.01 MB and 50.18 G multi-adds.  
Despite the parameter-heavy MLPs for CBV, the SP module remains highly efficient.  
Readers can decide if the MLPs are worth using based on specific applications.  
As shown in Fig.~\ref{fig:deformation_fields} (a) and (b), CBV produces a more regular and plausible displacement field without tuning the \(\lambda\)-parameter, improving TRE slightly from 1.93 to 1.88.
All baseline descriptor matching and learning-based methods, including initial alignment runtime, processed each scan pair in under one second.  

\begin{figure}[!t]
\centering
\captionsetup[sub]{font=scriptsize}
    \subcaptionbox{
    \label{fig:n_basis_crop}
    }{
        \includegraphics[width=0.234\columnwidth]{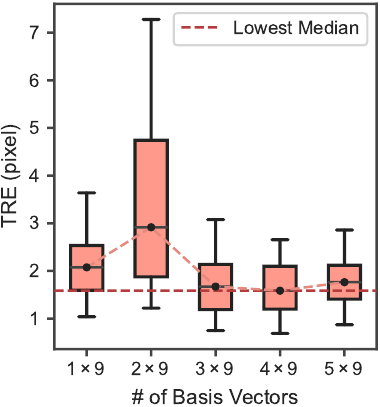}
    }
    \subcaptionbox{
    \label{fig:n_iters_crop}
    }{
        \includegraphics[width=0.342\columnwidth]{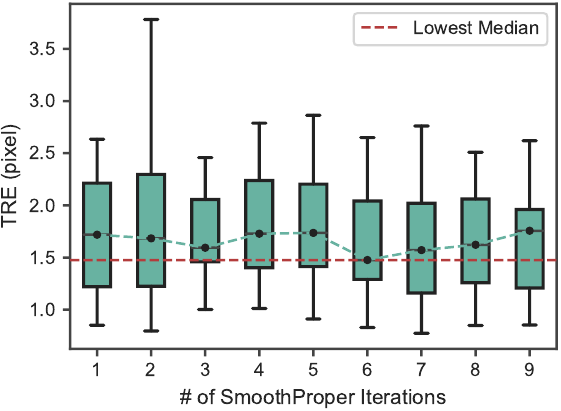}
    } 
    \subcaptionbox{
    \label{fig:img_size_crop}
    }{
        \includegraphics[width=0.252\columnwidth]{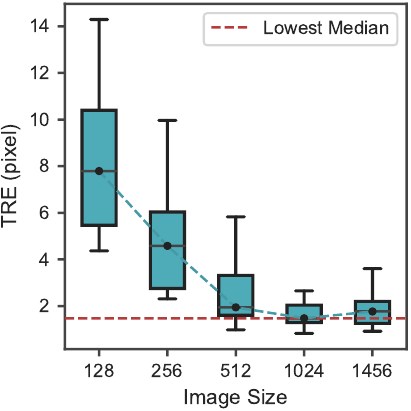}
    }
    \caption{
    Ablation analysis: boxplots and line plots showing the effect of basis vector count (a), inner-loop iterations (b), and input image size (c) on TRE (pixels).  
    Red dashed lines indicate the lowest median TRE in each plot.
    }
    \label{fig:ablation_fire}
\end{figure}

\textbf{Ablation Analysis on the Number of Basis Vectors \(m\):}  
In Fig.~\ref{fig:n_basis_crop}, we vary \(m\) from \(1 \times 3^2\) to \(5 \times 3^2\), where \(3^2\) mimics the local \(3 \times 3\) pixel neighborhood in discrete optimization.  
This neighborhood does not reflect actual displacements but corresponds to learned displacement atoms based on the local disparity context.
Larger \(m\) allows for richer information querying and better local structural consistency.  
However, increasing \(m\) too much can lead to limitations due to the available training samples.  
As shown, in FIRE, TRE decreases with \(m\) initially, reaching a minimum at \(4 \times 3^2\) before rising again.

\textbf{Ablation Analysis on Iterations \(K\):}  
We vary \(K\) from 1 to 9, as shown in Fig.~\ref{fig:n_iters_crop}, and find that the model achieves optimal performance at \(K = 6\).  
This result aligns with prior work \cite{zheng2015conditional} using mean-field inference \cite{krahenbuhl2011efficient} with 5 iterations, where further increasing \(K\) yields limited improvement.

\textbf{Ablation Analysis on Image Size:}  
We varied the squared input image size from 128 to 1024, doubling at each step, and included 1456 (half of the original size).  
As shown in Fig.~\ref{fig:img_size_crop}, 128 yields worse performance than the initial alignment.  
TRE decreases steadily from 256 to 1024 but increases at 1456.  
Smaller sizes improve efficiency but reduce accuracy, likely due to vessels becoming less visible.  
Larger sizes can introduce local noise, limiting accuracy gains.

%% file: docs/quant_fire.tex
\begin{table}[!t]
\centering
\caption{Quantitative comparison on the FIRE dataset.  
Best results are in bold.  
$\downarrow$ indicates lower is better, and $\uparrow$ indicates higher is better.  
The initial alignment method for learning-based approaches is highlighted in gray.
}
\resizebox{0.98\textwidth}{!}{%
\begin{tabular}{c||r||r|ccc|ccc}
\hline\hline
Category & Methods~~~~~~~~~~ & TRE $\downarrow$ & \multicolumn{3}{c|}{\textbf{AUC} 
 $\uparrow$} & \multicolumn{3}{c}{\textbf{Group-wise AUC@25} $\uparrow$} \\
\cline{4-9}
 &  &  & ~~mAUC@15~~ & ~~mAUC@25~~  & ~~mAUC@50~~ & ~mAUC@$\mathcal{A}$~ & ~mAUC@$\mathcal{P}$~ & ~mAUC@$\mathcal{S}$~ \\
\hline
\multirow{7}{*}{~~Detector-based Methods~~} 
 & XFeat \textcolor{gray}{\scriptsize[CVPR'24]} \cite{potje2024xfeat} & 10.858 & 0.560 & 0.637 & 0.794 & 0.853 & 0.102 & 0.915 \\
 & R2D2 \textcolor{gray}{\scriptsize [NeurIPS'19]} \cite{revaud2019r2d2} & 7.926 & 0.553 & 0.701 & 0.850 & 0.813 & 0.333 & 0.899 \\
 & LightGlue \textcolor{gray}{\scriptsize [ICCV'23]} \cite{lindenberger2023lightglue} & 7.802 & 0.575 & 0.710 & 0.855 & 0.853 & 0.338 & 0.904 \\
 & SuperPoint \textcolor{gray}{\scriptsize [CVPR'18]} \cite{detone2018superpoint} & 6.641 & 0.612 & 0.757 & 0.879 & 0.813 & 0.453 & 0.928 \\
 & Glampoints \textcolor{gray}{\scriptsize [ICCV'19]} \cite{truong2019glampoints} & 6.608 & 0.595 & 0.757 & 0.879 & 0.733 & 0.560 & 0.880 \\
 & SuperRetina \textcolor{gray}{\scriptsize [ECCV'22]} \cite{liu2022semi} & 6.382 & 0.622 & 0.767 & 0.884 & 0.813 & 0.516 & 0.909 \\
& \cellcolor{gray!20}RetinaIPA \textcolor{gray}{\scriptsize [MICCAI'24]} \cite{wang2024retinal} & \cellcolor{gray!20}5.750 & \cellcolor{gray!20}0.657 & \cellcolor{gray!20}0.774 & \cellcolor{gray!20}0.885 & \cellcolor{gray!20}0.799 & \cellcolor{gray!20}0.599 & \cellcolor{gray!20}0.886 \\

\hline
\multirow{5}{*}{~~Detector-free Methods~~} 
 & LoFTR \textcolor{gray}{\scriptsize [CVPR'21]} \cite{sun2021loftr} & 7.638 & 0.526 & 0.716 & 0.858 & 0.693 & 0.564 & 0.811 \\
 & DKM \textcolor{gray}{\scriptsize [CVPR'23]} \cite{edstedt2023dkm} & 6.493 & 0.610 & 0.760 & 0.880 & 0.800 & 0.529 & 0.891 \\
 & Aspanformer \textcolor{gray}{\scriptsize [ECCV'22]} \cite{chen2022aspanformer} & 6.415 & 0.602 & 0.761 & 0.881 & 0.800 & 0.556 & 0.877 \\
 & RoMa \textcolor{gray}{\scriptsize [CVPR'24]} \cite{edstedt2023roma} & 6.388 & 0.605 & 0.763 & 0.881 & 0.800 & 0.565 & 0.875 \\
  & GeoFormer \textcolor{gray}{\scriptsize [ICCV'23]} \cite{liu2023geometrized} & 6.201 & 0.625 & 0.770 & 0.887 & 0.813 & 0.587 & 0.925 \\

\hline
\multirow{3}{*}{\shortstack{Learning-based Methods \\ w/o Deep Unrolling}}
 & KeyMorph \textcolor{gray}{\scriptsize [MIDL'22]} \cite{evan2022keymorph} & 5.947 & 0.640 & 0.784 & 0.892 & 0.813 & 0.551 & 0.917 \\
 & C2FViT \textcolor{gray}{\scriptsize [CVPR'22]} \cite{mok2022affine} & 5.842 & 0.642 & 0.786 & 0.893 & 0.827 & 0.547 & 0.920 \\
 & GAMorph \textcolor{gray}{\scriptsize [BIBM'24]} \cite{liu2024progressive} & 3.081 & 0.825 & 0.895 & 0.945 & 0.906 & 0.800 & 0.949 \\
 \hline
\multirow{4}{*}{\shortstack{Learning-based Methods \\ w/ Deep Unrolling}}
 & GraDIRN \textcolor{gray}{\scriptsize [MICCAI'22]} \cite{qiu2022embedding} & 6.344 & 0.657 & 0.774 & 0.885 & 0.799 & 0.599 & 0.886 \\
 & PDD-Net \textcolor{gray}{\scriptsize [MICCAI'19]} \cite{heinrich2019closing} & 5.765 & 0.688 & 0.792 & 0.893 & 0.819 & 0.598 & 0.915 \\
& VR-Net \textcolor{gray}{\scriptsize [TMI'21]} \cite{jia2021learning} & 4.974 & 0.705 & 0.823 & 0.911 & 0.827 & 0.660 & 0.931 \\
 & \textbf{SmoothProper} (Ours) & \textbf{1.879}&\textbf{0.920} & \textbf{0.951} & \textbf{0.974} & \textbf{0.937} & \textbf{0.920} & \textbf{0.974}\\
\hline\hline
\end{tabular}%
}
\label{table:quant_fire}
\end{table}




%% file: docs/5_conclusion.tex


\section{Conclusions}

We introduce SmoothProper, a plug-and-play neural network module that enforces smoothness and message passing within the forward pass.
By integrating a duality-based optimization layer with tailored interaction terms, it efficiently propagates flow signals while preserving structural consistency.
SmoothProper addresses aperture and large displacement challenges in retinal vessel datasets, achieving a TRE of 1.88 pixels in unsupervised DIR for the first time.
Although this study focuses on the 2D problem, extending it to 3D is as simple as replacing the 2D convolution with a 3D one.
We hope our findings on utilizing the regularization term inspire new milestones and reignite interest in unsupervised DIR techniques.